\documentclass{bmvc2k}
\title{RecSal : Deep Recursive Supervision for Visual Saliency Prediction}

\addauthor{\hspace{15mm} Sandeep Mishra* \hspace{40mm} Oindrila Saha*}{}{1}
\addauthor{\hspace{31mm} Indian Institute of Technology Kharagpur}{}{1}

\addinstitution{}

\runninghead{Mishra, Saha}{RecSal}

\def\etal{\emph{et al}\bmvaOneDot}

\begin{document}

\maketitle

\begin{abstract}
State-of-the-art saliency prediction methods develop upon model architectures or loss functions; while training to generate one target saliency map. However, publicly available saliency prediction datasets can be utilized to create more information for each stimulus than just a final aggregate saliency map. This information when utilized in a biologically inspired fashion can contribute in better prediction performance without the use of models with huge number of parameters. In this light, we propose to extract and use the statistics of (a) region specific saliency and (b) temporal order of fixations, to provide additional context to our network. We show that extra supervision using spatially or temporally sequenced fixations results in achieving better performance in saliency prediction. Further, we also design novel architectures for utilizing this extra information and show that it achieves superior performance over a base model which is devoid of extra supervision. We show that our best method outperforms previous state-of-the-art methods with 50-80\% fewer parameters. We also show that our models perform consistently well across all evaluation metrics unlike prior methods.   
\end{abstract}

%-------------------------------------------------------------------------
\vspace{-18pt}

\begin{wrapfigure}{r}{3cm}
\includegraphics[width=3cm]{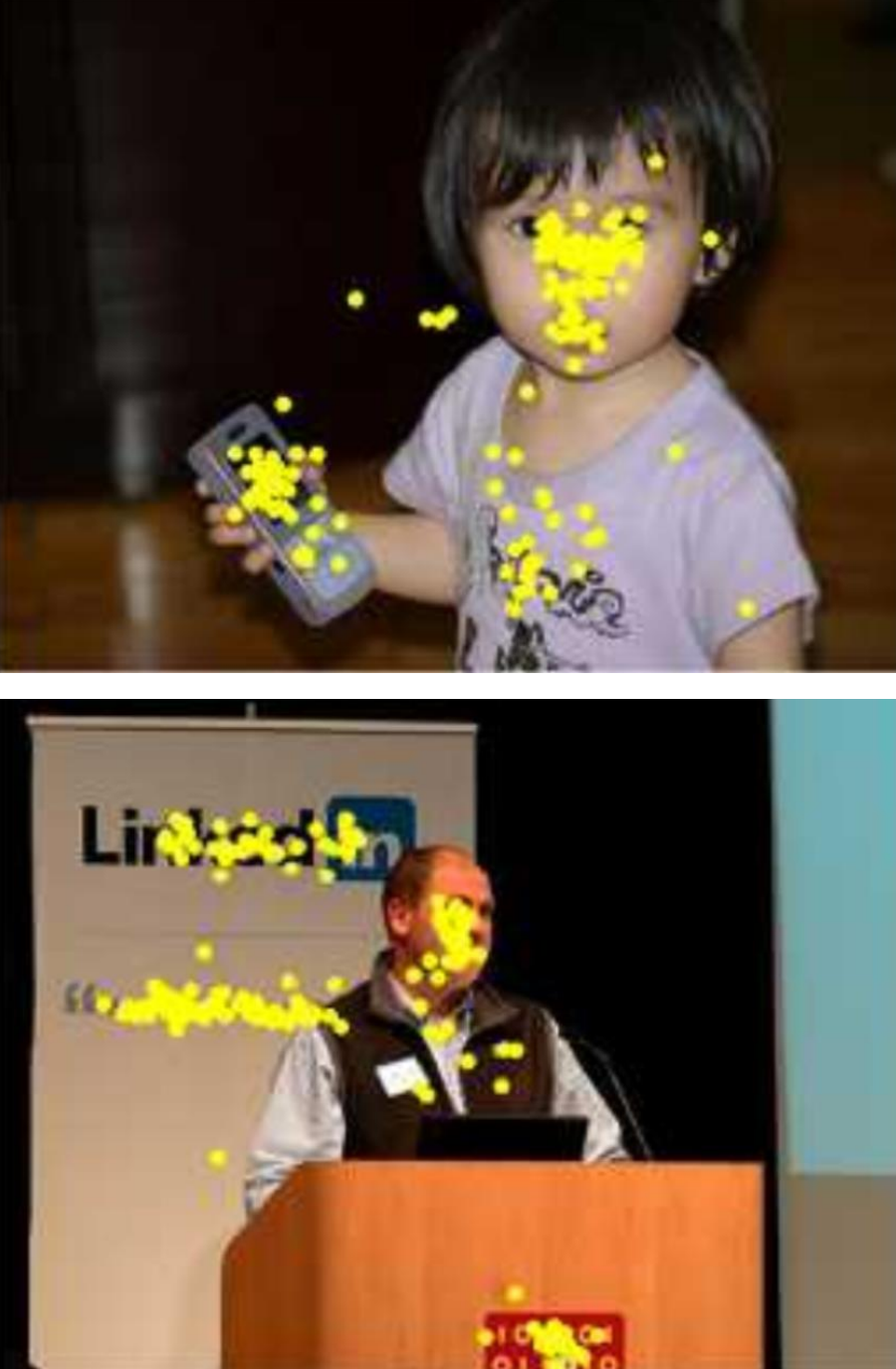}
\vspace{-15pt}
\caption{\footnotesize Some example OSIE\cite{xu2014predicting} images with fixation points in yellow}\label{fig:osie}
\end{wrapfigure}

%\vspace{-10pt}

\section{Introduction}
\label{sec:intro}
Visual saliency is the probability of spatial locations in an image to attract human attention. Given an image, mimicking human saliency patterns is a key to solve many vision problems. To enable this, saliency prediction models must be presented with data in a similar manner as to humans. Essentially, when presented with an image a human subject will look at locations one-by-one with their next fixation depending on what they have already seen. Further, when a person looks at a particular region there will be a pattern as to how that region is scanned i.e. what are the most interesting points in that region. Thus the information of temporal sequence as well as region-specific saliency patterns is important to be recognized by a model to predict a saliency map imitating human fixation probabilities. 

Data-driven approaches to saliency prediction depend upon the ground truth aggregate saliency map to train deep CNN models end-to-end. However, such a map is a crude average over observers, time and spatial regions. The saliency of an image can be very different across regions and time. Once an observer is presented with an image, they can start with any point in the image, then move on to fixate at any another location. Now the next location that a person fixates on will be dependent on the location he has already scanned. So a fixation at a given time is dependent on all the fixations of the observer before that \cite{itti2006bayesian}. This path of all fixations arranged wrt time is known as a scanpath \cite{noton1971scanpaths}. The aggregate saliency map does not contain this temporal scanpath information. But this information is crucial in predicting saliency due to the successive dependency.

Since typical saliency prediction models predict a single map, fixation points in all spatial regions of the image are treated in the same way. Let us contrast this with a segmentation network where each object is assigned a different class channel in the final model output, resulting in the model being able to learn to treat separate regions differently. But in saliency predicition, irrespective of regions or objects, the ground truth is just a single map. It should also be noted that the fixation pattern varies wrt the types of objects in a region. Humans look at different regions in the image in different manner. For example, in the images in Figure \ref{fig:osie} a face tends to have fixations where the features lie, like eyes, nose and mouth. The phone in the child's hand has fixations on the screen, while uniform spaces like the background of the man in the bottom image has fixations on the texts. Treating these areas differently and separately learning them can allow the network to learn local saliency patterns.

Our prime contributions are as follows:
\vspace{-4pt}

\begin{itemize}
    \itemsep 0pt
    \topsep 0pt
    \parskip 2pt
    \item We propose a multi-decoder network to exploit the contributions of features obtained from shallow and deep layers of the encoder to form the final saliency map.
    \item We design our model to predict multiple saliency maps each of which is trained on a separate loss so as to enable one model to do well in all evaluation metrics.
    \item We propose the use of temporally and spatially sequenced metadata to provide bio-inspired deep supervision to our model. To the best of our knowledge, this is the first method to use such extra data for supervision in a saliency prediction task.
    \item We further propose novel recursive model architectures to effectively use this metadata for sequential supervision, and finally show superior results in predicting the aggregate saliency maps, as compared to our baselines and previous state-of-the-art methods.
\end{itemize}

% \vspace{-8pt}

% \vspace{5pt}
\section{Related Work}
\label{sec:related}

\textbf{Visual Saliency:} Since the classical methods \cite{itti1998model, treisman1980feature, harel2007graph} which utilized hand-crafted features, saliency prediction models have come much closer to mimicking humans with the help of deep CNNs. Advances in model architectures have shown to obtain better performance. SalGAN \cite{pan2017salgan} uses an added adversarial network for loss propagation. Dodge \etal \cite{dodge2018visual} uses two parallel networks to fuse feature maps before prediction. We propose a novel base architecture with multiple decoders which utilize features extracted from coarse to finer levels for predicting saliency. Performance is measured using varied metrics and it has been shown that optimizing a model on a single one of these will not give good performance on the other metrics. Kummerer \etal \cite{Kummerer_2018_ECCV} states that a probabilistic map output which can be post-processed for various metrics tackles this problem. Some methods incorporate more than one loss to train the final saliency output \cite{jia2020eml,cornia2018predicting}. We design our model to produce multiple output saliency maps each optimized on a different metric.                                                                                                                                                                                                       \par                                                                                                          \noindent\textbf{Recursive Feature Extraction:} Some methods \cite{cornia2018predicting, wang2018salient} use a recurrent module for refining the final saliency map recursively. Fosco \etal \cite{fosco2019many} uses multi duration data and uses a variant of LSTM \cite{hochreiter1997long} based attention module to compute outputs denoting where people look for a given time duration after being shown the image. Jiang \etal \cite{jiang2018deepvs} uses a Convolutional LSTM \cite{xingjian2015convolutional} based method for video saliency prediction. We use a recursive block (\(RB\)) to provide extra auxilliary supervision using sequential data to finally improve saliency prediction performance.

\par

\noindent\textbf{Additional Context Using Extra Data:} Approaches like \cite{islam2018semantics, li2016deepsaliency} perform multi-task learning to solve both segmentation and saliency prediction, and show how the added context affects each task. Some methods \cite{dmitriev2019learning, saha2019learning} use information from multiple datasets to solve an unified task -- segmentation in this case. Zhao \etal \cite{zhao2015saliency} provides local and global context using the same input image with the help of two independent networks. Ramanishka \etal \cite{ramanishka2017top} uses caption generation in videos for performing more accurate saliency prediction. As opposed to these methods, we do not use any extra dataset or annotations for auxillary supervision. We describe how we extract this data from eye gaze annotations in Section \ref{sec:extsup}. We use temporally and spatially sequenced data for a single static image to provide extra context to the network to enhance performance.

\begin{figure*}
% \begin{center}
% \fbox{\rule{0pt}{2in} \rule{.9\linewidth}{0pt}}
% \end{center}

\includegraphics[width=\linewidth]{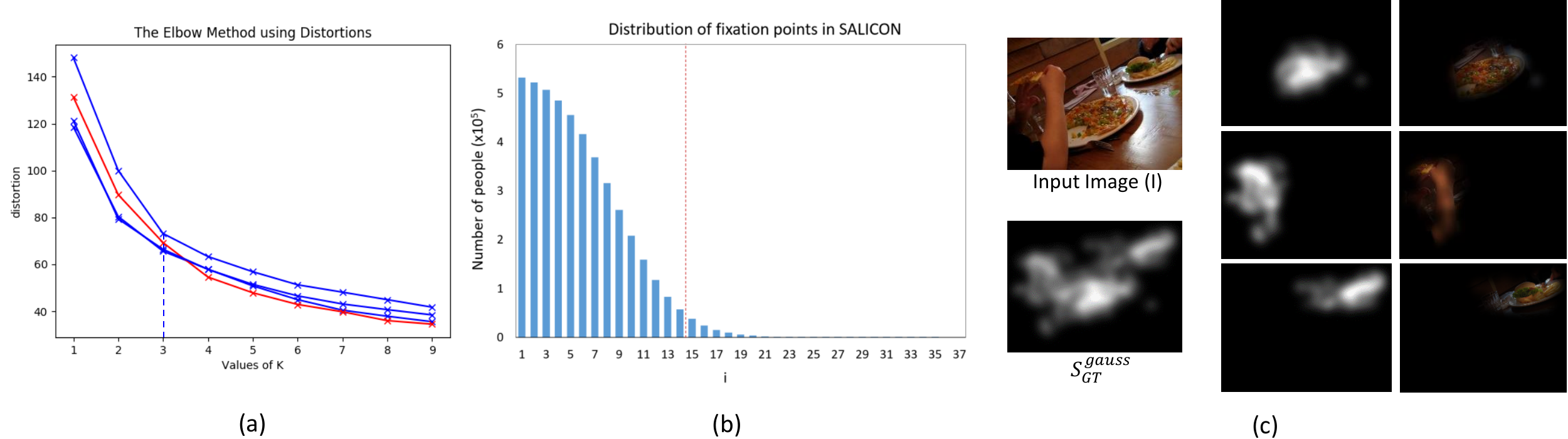}
   \caption{\footnotesize(a) Plot of Distortion vs number of clusters (K) where the usual elbow point is at 3. The red line highlights an exception where the elbow point is 4.  (b) Plot of number of people who have at least '\(i\)' fixations vs '\(i\)' for SALICON train set. (c) For given input image and ground truth \(S^{gauss}_{GT}\), the region based separated maps ordered wrt the number of fixation points present in each cluster. The 3 salient regions visible: pizza, hand and burger are separated by our algorithm. The rightmost column is an overlay of the three maps on the image, which shows that the pizza is of highest interest, followed by the hand and then the burger.}
\label{fig:data}
\end{figure*}

% \vspace{1pt}

\section{Extra Supervision}
\label{sec:extsup}

We use SALICON-2017 \cite{huang2015salicon} and MIT1003 \cite{judd2009learning} datasets for training our models. Rather than using extra annotations and enforcing the network to learn how to solve a divergent task, such as semantic segmentation \cite{islam2018semantics}, we propose to use data that is directly derived from eye gaze as that would provide more task specific context to the network. We create this meta-data using information already available in the eye gaze annotations of the above datasets to add deep supervision to our network. We separate and order aggregate fixation points based on spatial region-specific importance and temporal sequence as explained below.

% \cite{them} use semantic segmentation maps as supervision on top of the traditional method of providing loss of the predicted saliency map from the target map. But this method makes the task of saliency prediction difficult for the network rather than making it easy because now the network also has to learn to segment the image into all the different 
% classes on which it is trained. Similarly \cite{them} use captions provided for PASCAL VOC while training for saliency prediction, which again is a deviation since the network has to learn to understand the image based on the caption which has been collected separately and might not have very high contextual information. To mitigate this distance we propose to create meta-data from the SALICON or MIT saliency prediction dataset itself, so that we can capture as much as contextual information as possible and make it easier for the network to learn saliency prediction.

\subsection{Temporal Data}
\label{sec:temporal}
Both the SALICON and MIT1003 provide eye gaze data which contains fixation points in the order of the occurrence of those fixations, hence we choose to provide this additional temporal information to the network. This will enable the network to extract information about the image as humans do, i.e. first on a coarser level and then go on to perceive the finer details as the image is seen for more time. 

MIT1003 uses the second to sixth fixation point of all users to generate the final fixation map per image. They choose to ignore the first fixation for each user to avoid an extra point because of the initial centre bias. We separate these points based on order of occurrence wherein for a particular image we create five temporally sequenced fixation maps. Each map contains the \(i^{th}\) fixation point for all viewers who saw that image. However, SALICON uses all the fixation points for each user to form the final saliency map. On careful examination of the data we find that there were images which had as less as zero or one fixation and as high as 35 fixations per user. Therefore, we plot a histogram of the number of people having at least \(i\) fixations (Figure \ref{fig:data} (b)).We observe that the histogram follows an approximately gaussian distribution with the peak being at 1. We find the standard deviation ($\sigma$) of our curve to be ~6.7 and using the '68-95-99.7' / '3-sigma' rule we choose the $\mu + 2\sigma$ point, before which 95\% of the information lies, to be our number of temporally sequenced fixation maps. Thus the number of maps become fourteen, which we make as described for MIT1003 and put the rest all remaining fixations in one last map. So our total maps and thus timesteps here become fifteen. Note that we ignore the first fixation point of every user for MIT1003 but keep them for SALICON so that the aggregate of temporally sequenced fixation maps align with the ground truth fixation map provided in the datasets.

\vspace{-6pt}
\subsubsection{Temporal Order vs Duration}
\label{sec:ordervsdur}
The temporally sequenced fixation maps generated above can be used in two ways. One is to simply arrange these maps in order of occurrence as the output target for each time-step of our recurrent module. The second way is to modify these maps as \(m_t = m_{t-1} + m_{t-2} + \dots + m_0\). This means that the \(t^{th}\) map consists of all the locations that have been looked at till the \(t^{th}\) time-step i.e. in a given \textbf{duration} from 0 to \(t\). We denote the maps arranged in temporal order as non-incremental data and the duration-wise arranged maps as incremental data.
\vspace{-5pt}
\subsection{Spatial Data}
\label{sec:spatial}
Close observation of the gaze data suggests that people tend to focus more on certain regions than just hovering through the whole image which tells us that there are certain spatial locations which are regions of interest and hence attracting viewers' gaze. Thus, given an image stimulus we enable the model to identify which regions are more salient and sequence them in order of importance to help form the final saliency map better. This is just the mapping of the relative importance of these regions in a single prediction.

Given the aggregate fixation points of all users across time, we use Gaussian Mixture Models \cite{reynolds2009gaussian} over the 2D point maps to create clusters. We use the elbow method on the WSS (within-cluster sum of square) vs number of clusters plot for each map of both SALICON and MIT1003 and find that the optimal number of clusters is three (Figure \ref{fig:data} (a)). Thus, we divide the fixation points into five sets and order them with respect to the total points in each set, more points in a set denoting spatial regions of higher interest. 

Note that the same technique as described in Section \ref{sec:ordervsdur}, can be performed for the spatially sequenced saliency maps also. We use the terminology of \textbf{incremental} vs \textbf{non-incremental} similarly as temporal data for spatial data as well. Note that we will denote the incremental metadata maps as \(M^{I} : \{m^{I}_0, m^{I}_1, \dots, m^{I}_{T-1}\}\) (as the last map \(m^{I}_{T}\) is same as the final saliency map \(S\)); and the non-incremental maps as \(M^{NI} : \{m^{NI}_0, m^{NI}_1, \dots, m^{NI}_T\}\) from now onwards. We compare the effect of above methods of deep supervision in Section \ref{sec:experiments}. 

\begin{figure*}
% \begin{center}
% \fbox{\rule{0pt}{2in} \rule{.9\linewidth}{0pt}}
% \end{center}
\includegraphics[width=\linewidth]{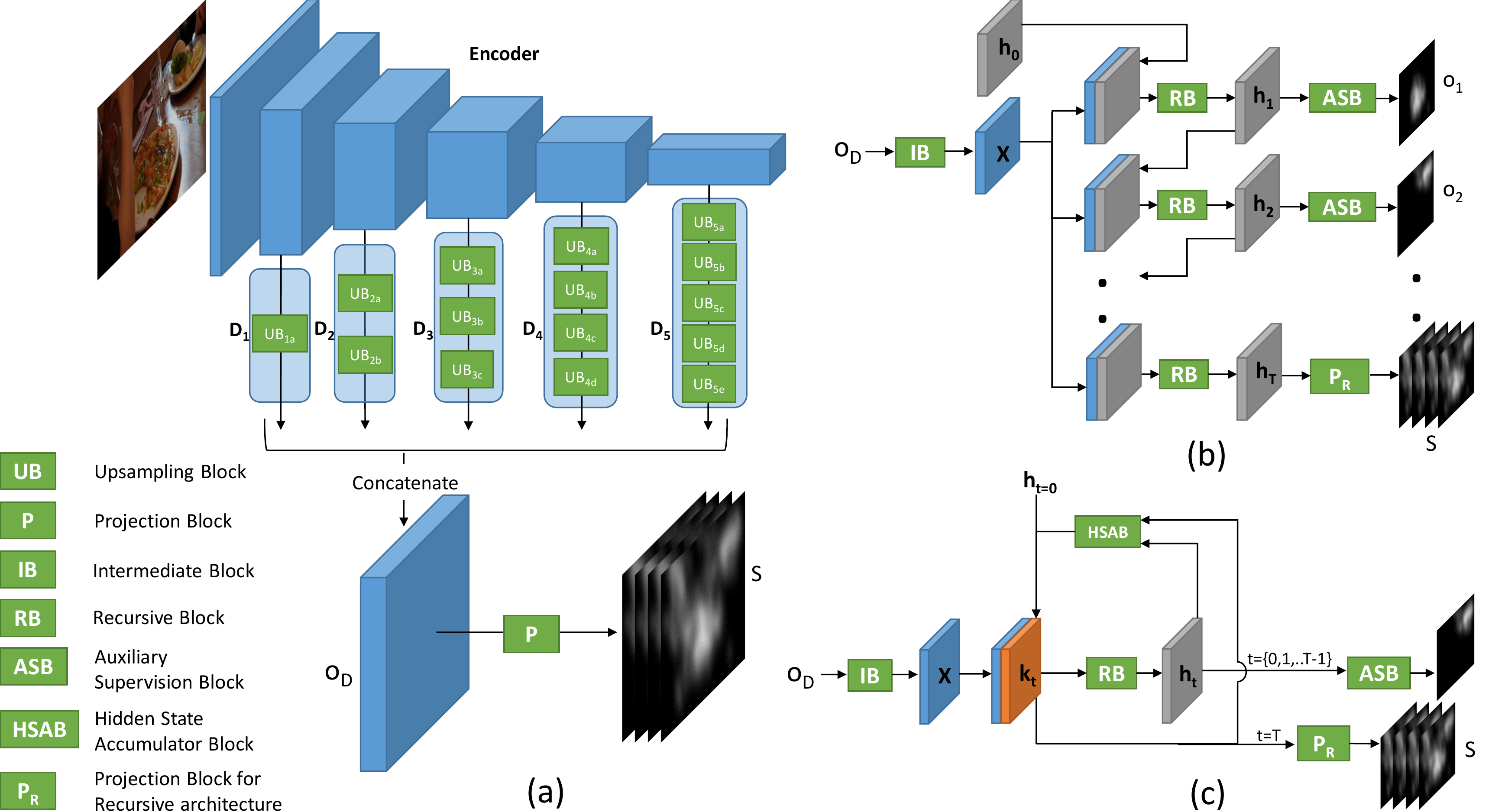}
\vspace{-5pt}
   \caption{\footnotesize(a) Base encoder-multi decoder architecture, (b) Recursive Module for Incremental Data (\(RB\)), (c) Recursive Module for Non-Incremental Data built upon \(RB\) with the addition of \(HSAB\) module}
\label{fig:arch}
\end{figure*}

\vspace{-10pt}

\section{Model Architecture}
\label{sec:arch}
\vspace{-5pt}
\subsection{Base Architecture}
\label{sec:basearch}
We design a deep convolutional encoder decoder network for our base architecture. As described in Itti \etal \cite{itti1998model}, features like color, intensity and textures contribute in determining saliency. Hence we choose to use feature maps from the shallowest to the deepest levels of the encoder for final saliency prediction. We do so using multiple decoders which consume feature inputs from the encoder just before every downsampling operation. The first decoder is placed just before the second downsampler and we use a total of five decoder blocks which gives us the advantage of using features from five different scales.

For the decoders we use stacks of up-sampling blocks, which are bilinear upsampling operations followed by a convolution + batch norm \cite{ioffe2015batch} + ReLU \cite{agarap2018deep} layer, such that final outputs of each decoder has same spatial size. Finally the outputs of all decoders are concatenated to form \(o_D\) (where \(D : [D_1, D_2, D_3, D_4, D_5]\)) which is passed through the projection convolution block: \(P\) to form the final output multiple saliency maps as shown in Figure \ref{fig:arch} (a). Note that each output map optimises on a different loss as explained in Section \ref{sec:losses}.

\vspace{-4pt}

\subsection{Recurrent Module for Incremental Data}
\label{sec:recurrentinc}
We need to modify the above base architecture for effectively using the metadata (\(M\)) described in Section \ref{sec:extsup}. We first consider using the incremental metadata. Since this extra data can directly be used to construct the final saliency ground truth map without extensive operations, one option is to have an intermediate activation map: \(X\) from the projection convolution block (\(P\)) trained to predict these generated maps \(m_t\) via an auxiliary loss: \(L_{aux}(X, M)\) while still retaining the output of \(P\) to be \(S\) (the final aggregate saliency maps). But such an architecture will not be able to exploit the relation between each \(m_t\) effectively, as the channel order does not affect the learning of the model. Therefore, to use the sequential data meaningfully, we shall have to predict each \(m_t\) successively. But \(X\) itself is not self-sufficient to create a sequential mapping between each \(m_t\). As a result we will need a recursive formulation where each \(m_t\) will be predicted with the help of \(X\) and a hidden state \(h_{t-1}\) which encodes all information till \(t-1\). Using an initial state of no information (\(h_0\)) and \(X\), we can create the first hidden state \(h_1\) which subsequently when passed through an Auxiliary Supervision Block: \(ASB\) will generate \(o_1\) (see Figure \ref{fig:arch}). Now this map \(h_1\) along with \(X\) can be used to generate \(h_2\), hence forming a recurrence relation as below where \(h_t\) contains all the information up to time step \(t\) (here \(f^{IC}_{1}\) is \(RB\) and \(f_{aux}\) is \(ASB\)): 

\vspace{-8pt}

% which can be done by using few layers of convolution on the intermediate activation map: \(X\) in (\(P\)) .

\begin{align}
\label{1}
h_{t} = f^{IC}_{1}(X,h_{t-1});
\quad\quad\quad o_{t} = f_{aux}(h_t);
\end{align}

Once we generate \(o_t\) \(\forall\)  \(t\), we concatenate them in order of \(t\) to form \(O\) (where \(O : [o_1, o_2, \dots, o_{T-1}]\)) and use it to provide \(L_{aux}(O, M)\) for supervision. Here we exclude \(o_{t=T}\) because this map contains decoded information about all time steps from \(0\) to \(t\). We send this map to the Projection block  (\(P_R\)) which then generates the final saliency map \(S\). To implement \eqref{1} we need to design a recursive unit which will be able to encode the dependency sequentially. Now, since spatial information is crucial to determining all of \(m_t\) or \(S\), it is clear that we cannot use the vanilla fully connected RNN or LSTM architectures, but need to use convolution operations on our activation maps. \cite{xingjian2015convolutional} uses convolution operations on the input and hidden state spatial maps. However, the input to our recurrent block is always a fixed (\(X\)) for all \(t\), which means that determining the next \(h_t\) will only be dependant on hidden state (\(h_{t-1}\)) and cell state (\(c_{t-1}\)) according to the equations of ConvLSTM. Thus, we need to encode a more complex relationship between \(X\) and \(h_{t-1}\) rather than linear addition to produce \(h_{t}\). To incorporate this, we propose a Recurrent Convolutional Block (\(RB\)) which is constructed by stacking three convolution layers each followed by a batchnorm and ReLU. Figure \ref{fig:arch} (b) shows the recursive block used in our model. We provide \(X\) and \(h_{t-1}\) concatenated as input to this block, which then gives us \(h_t\). This can be formulated as a non-linear function on both \(X\) and \(h_{t-1}\) i.e. \(h_t = f(X,h_{t-1})\). But, if we have a decouple \(W\) and \(U\) as suggested by ConvLSTM, then the relation becomes \(h_t = f_W(X) + f_U(h_{t-1})\). Note that all \(f\), \(f_W\) and \(f_U\) have more than one convolution layers each followed by batchnorm and ReLU; and thus are non-linear functions. Since \(f\) is a function over the combined space of \(X\) and \(h_{t-1}\) it can learn much more complex relation among the two rather than a linear sum of independent functions applied on them.

% \vspace{-4pt}

\subsection{Recurrent Module for Non-Incremental Data}
\label{sec:recurrentnon}

The above formulation will not help us in our non-incrementally arranged data. Now the metadata maps \(m_t\) will only contain information of the time step \(t\) and not all the information up to it. Since \(h_t\) is optimized to predict \(m_t\) as we apply the supervision loss \(L_{aux}\) on the output of \(ASB(h_t)\); it tends to have less information about all the previous states while focusing only on the current state. This makes it much more difficult for \(RB\) to predict \(h_t\) from \(h_{t-1}\). Assuming \(RB\) can learn to output \(h_t\) which contains information about time step \(t\) only, the next input to \(RB\) then cannot be the same \(h_t\) since it does not contain all the information up to time \(t\) which is necessary for the sequence to be generated in order. Hence, we introduce a Hidden State Accumulator Block (\(HSAB\)), which keeps track of all the hidden states up to time \(t\) and generates an accumulated output \(k_t\) which now contains all the information from time \(0\) to time \(t\). Now that we have modified \(RB\) to generate \(h_t\) \(\forall\) \(t\), we pass each of them through \(ASB\) and then concatenate all the outputs to form \(O\) (where \(O : [o_1, o_2, \dots, o_{T}]\)). This \(O\) is then used to compute the supervision loss \(L_{aux}(O, M)\). The last \(h_T\) is again passed through \(HSAB\) to get \(k_T\) which can now be passed through \(P_R\) (see Figure \ref{fig:arch}) to get the final saliency maps since \(k_T\) contains all the information up to the last time instant T. This can be expressed as follows where \(f^{NIC}_{1}\) is \(RB\), \(f^{NIC}_{2}\) is \(HSAB\) and \(f_{aux}\) is \(ASB\).

% \vspace{-4pt}

\begin{equation}
\label{2}
h_{t} = f^{NIC}_{1}(X,k_{t-1});
\quad\quad\quad k_{t} = f^{NIC}_{2}(h_{t},k_{t-1});
\quad\quad\quad o_{t} = f_{aux}(h_t) \hspace{1pt} \forall \hspace{1pt}t;
\end{equation}

\section{Losses}
\label{sec:losses}
Recent works suggest using saliency evaluation metrics like \(KL\) (Kullback–Leibler divergence), \(SIM\) (Similarity), \(CC\) (Pearson's Correlation Coefficient) and \(NSS\) (Normalized Scanpath Saliency) as losses during training. According to \cite{riche2013saliency} and \cite{Kummerer_2018_ECCV}, metrics for comparison of saliency are not coherent i.e. every metric penalizes different aspects in the saliency map. So, training with a single map optimizing on all these metrics \cite{jia2020eml,cornia2018predicting,fosco2019many} will not be able to bring out the best performance of the model for each score. Thus, we propose to have multiple saliency map outputs from our model so as to optimize each map on a different loss. \cite{riche2013saliency} shows how different each metric is from each other, thus to optimize our model on most metrics we choose \(KL\), \(CC\), \(SIM\) and \(NSS\) to train our network. Since all kinds of \(AUCs\) are not differentiable, we exclude them in our losses. We use four output saliency maps such that our total loss is as follows. 
\vspace{-3pt}

\begin{equation}
\label{3}
\small L_{sal}(S, S_{GT}) = \alpha L_{KL}(S_1, S^{gauss}_{GT}) + \beta L_{CC}(S_2, S^{gauss}_{GT}) + \gamma L_{SIM}(S_3, S^{gauss}_{GT}) + \delta L_{NSS}(S_4, S^{pts}_{GT})
\end{equation}

Here, \(L_{KL}\) is the standard \(KL\) loss as defined in \cite{bylinskii2018different}, \(L_{CC}\) is \(1-CC\), \(L_{SIM}\) is \(1-SIM\) and \(L_{NSS}\) is \(-NSS\). Note that \(S^{pts}_{GT}\) is the ground truth map with fixation points, while \(S^{gauss}_{GT}\) is the \(S^{pts}_{GT}\) blurred using antonio gaussian kernel as in  \cite{judd2009learning}. The values of $\alpha$, $\beta$, $\gamma$ and $\delta$ are chosen after experimentation as described in Section \ref{sec:experiments}.

As our metadata maps, we can have both \(M^{pts}_{GT}\) and the blurred \(M^{gauss}_{GT}\). However, the effects of the antonio gaussian function depends on the spatial distribution of points in the binary fixation map. This means that now when a binary fixation map is broken down into \(m^{fix}_{0\le t\le T}\) maps, the spatial distribution of points change drastically with respect to the original fixation map. As a result in each \(m^{gauss}_{0\le t\le T}\) obtained from corresponding \(m^{fix}_{0\le t\le T}\), the intensity values at a given region (where fixation points are present in \(m^{pts}_{t}\)) are different from that of the same region in \(S^{gauss}_{GT}\). So, using \(M^{gauss}_{GT}\) would make learning the final map \(S\) harder. Therefore, since only \(NSS\) uses the fixation points for error calculation we use only \(M^{pts}_{GT}\) for supervision and hence \(L_{aux}(O,M) = \frac{1}{T} \sum_{t=0}^{T} L_{NSS}(o_{t}, m^{pts}_{t})\).

% \vspace{-2pt}

\section{Experiments} 
\label{sec:experiments}

For empirical evaluation, the proposed network architectures are trained and tested on 3 publicly available datasets. We conduct a detailed ablation study to find our best performing model settings. Thereafter, we use this model to compare against state-of-the-art methods.

\textbf{Dataset: }Commonly used datasets like MIT1003 \cite{judd2009learning} and OSIE \cite{xu2014predicting} are not sufficient enough for training huge models with millions of parameters. We use the MIT1003 dataset which uses gaze tracking devices on 15 subjects per image and total of 1003 images to create their dataset. Another similar dataset is the OSIE dataset which contains eye tracking data on 700 images. While the images in these datasets cover a variety of scenes, the number of images isn't sufficient enough to train deep models. Therefore, for training purpose we use SALICON \cite{huang2015salicon} which has 10000 training and 5000 validation images with well defined target saliency maps and also accepts submissions in their online competition named Large Scale Scene-Understanding (LSUN 2015 and 2017). We perform all our experiments on the 2017 data. For this competition they provide 5000 test images and the results are to be submitted at the given website. SALICON images have a consistent size of 480x640 and they use mouse tracking data to create the corresponding annotations. For the other datasets, we perform fine-tuning on the model trained on SALICON with their respective train sets.

% SALICON has proved to perform consistently when compared with datasets like OSIE and MIT1003. Hence the community has started using this dataset to pre-train their model on this dataset and then fine tune on other smaller datasets, which is why we opted to do the same. 

% \vspace{-8pt}

\textbf{Evaluation Metrics: }Previous literature suggests various metrics for evaluating saliency prediction and it is general convention to provide results on many of them for fair comparison since each metric has its own way of measuring performance. We use \(NSS\), \(KL\), \(CC\), and \(SIM\) to compute our validation scores and to evaluate ablation study, while our results on testing dataset of SALICON are also evaluated on \(sAUC\), \(AUC\textunderscore judd\), \(AUC\textunderscore borji\) and \(IG\). For a detailed discussion on the definition and properties of the mentioned metrics please refer \cite{bylinskii2018different}. 

\textbf{Training Methodology: }We perform initial experiments using the base architecture structure (Figure \ref{fig:arch} a) alongwith multi-channel output. First we chose ResNet18 \cite{he2016deep} pre-trained on ImageNet1K \cite{deng2009imagenet} as our encoder network and trained the model with four output maps using the losses mentioned in Section \ref{sec:losses}. We train for 10 epochs with cosine learning rate scheduler \cite{loshchilov2016sgdr} wherein we start with a optimal lr of 1e-3, with a batch size of 35 and use SGD optimizer. Since we train on SALICON which has consistent image sizes, hence we choose not to resize the image and train the network with full size image to avoid losing information during resizing. The parameters used in our loss function were found to be $\alpha = 2$, $\beta = 2$, $\gamma = 5$ and $\delta = 1$ after an extensive search. These parameters were chosen so as to obtain good results on all the metrics.

% We chose these values based on the contribution of each loss value at convergence. 

% We also did a brief search around these values of $\alpha$, $\beta$, $\gamma$ and $\delta$ and found that the change in validation scores were inconsiderable.

On achieving best possible validation scores with ResNet18 we shifted to DenseNet121 \cite{huang2017densely} to observe effects of the accuracy of ImageNet pretraining affecting saliency prediction. Even though ResNet18 has almost double parameters of DenseNet121, it still performs worse on ImageNet classification. Results on comparison of the two encoders are shown in Table \ref{ablation}. Since DenseNet121 clearly has a much superior performance than ResNet18 when used in our setup, we use it in all our other experiments. We choose not to use any bigger encoder than DenseNet121 to avoid any further increase in number of parameters. Note that we use encoders pre-trained on ImageNet1K following prior art methods for fair comparison of performance. 
 
\textbf{Training with Metadata: }After evaluating the best base model we move to investigate our recursive model based on usage of temporal metadata. We train using both the incremental and non-incremental data in their corresponding architectures and evaluate. We observe that training takes slightly more time to converge than the base architecture. Hyperparameters were kept the same as the base architecture to ensure fair comparison and the auxiliary loss \(L_{aux}\) was given a weight of 0.01 after extensive search. Similar experiments were performed for spatial metadata as well. All the results of the ablation study for these various settings are recorded in Table \ref{ablation}. We observe that the recursive model trained on non-incrementally arranged spatial metadata performs the best among all the variations. Hereafter, we compare this \textbf{non-incremental spatial} model - \textbf{RecSal-NIS} with prior state-of-the-art methods on various datasets. The results of comparison on MIT1003 and SALICON validation sets are recorded in Table \ref{MITval} and Table \ref{SALICONval}. Note that all the other architectures in the prior art use much heavier encoders like ResNet-50, VGG-16 and DenseNet-161 which have close to 100M parameters as compared to our 15.56M. This shows that our method achieves competitive performance in much fewer parameters.

\begin{table}[]
\center
\resizebox{1\columnwidth}{!}{
\begin{tabular}{@{}l|l|l|cccc|cccc@{}}
\hline
{ } &
  { } &
  { } &
  \multicolumn{4}{c|}{{ \textbf{SALICON}}} &
  \multicolumn{4}{c}{{ \textbf{MIT1003}}} \\
\multirow{-2}{*}{{ \textbf{Training data}}} &
  \multirow{-2}{*}{{ \textbf{Architecture}}} &
  \multirow{-2}{*}{{ \textbf{Procedure}}} &
  { \textbf{KL}} &
  { \textbf{CC}} &
  { \textbf{SIM}} &
  { \textbf{NSS}} &
  { \textbf{KL}} &
  { \textbf{CC}} &
  { \textbf{SIM}} &
  { \textbf{NSS}} \\ \hline
{ } &
  { ResNet18 + D + \(P_R\)} &
  { -} &
  { 0.239} &
  { 0.876} &
  { 0.743} &
  { 1.983} &
  { 0.71} &
  { 0.723} &
  { 0.542} &
  { 2.872} \\
\multirow{-2}{*}{{ SALICON}} &
  { DenseNet121 + D + \(P_R\)} &
  { -} &
  { 0.224} &
  { 0.887} &
  { 0.761} &
  { 1.998} &
  { 0.698} &
  { 0.747} &
  { 0.551} &
  { 2.941} \\ \hline
{ } &
  { DenseNet121 + D + \(RB\) + \(P_R\) + ASB} &
  { Incremental} &
  { 0.215} &
  { 0.891} &
  { 0.786} &
  { 2.009} &
  { 0.687} &
  { 0.756} &
  { 0.569} &
  { 3.032} \\
\multirow{-2}{*}{{ SALICON + Temporal MetaData}} &
  { DenseNet121 + D + \(RB\) + HSAB + \(P_R\) + ASB} &
  { Non-Incremental} &
  { 0.219} &
  { 0.894} &
  { 0.792} &
  { 2.016} &
  { 0.685} &
  { 0.761} &
  { 0.56} &
  { 3.035} \\ \hline
{ } &
  { DenseNet121 + D + \(RB\) + \(P_R\) + ASB} &
  { Incremental} &
  { 0.215} &
  { 0.901} &
  { 0.792} &
  { 2.014} &
  { 0.672} &
  { 0.781} &
  { 0.576} &
  { 3.051} \\
\multirow{-2}{*}{{ SALICON + Spatial MetaData}} &
  { DenseNet121 + D + \(RB\) + HSAB + \(P_R\) + ASB} &
  { Non-Incremental} &
  { \textbf{0.206}} &
  { \textbf{0.907}} &
  { \textbf{0.803}} &
  { \textbf{2.027}} &
  { \textbf{0.665}} &
  { \textbf{0.784}} &
  { \textbf{0.583}} &
  { \textbf{3.074}} \\ \hline
\end{tabular}}
\vspace{-7pt}
\caption{\footnotesize Ablation study over architectures and metadata types for validation sets of SALICON and MIT1003}
\label{ablation}
\end{table}

%%%%% do wala table
\begin{table}
\centering
% \hspace{0.5cm}
\parbox{.49\linewidth}{\resizebox{0.49\columnwidth}{!}{
\centering
\begin{tabular}{l|cccc}
\hline
{ }        & \multicolumn{4}{c}{{ \textbf{SALICON}}}                                                               \\
\multirow{-2}{*}{{ \textbf{Method}}} &
  { \textbf{KL} \(\downarrow\)} &
  { \textbf{CC} \(\uparrow\)} &
  { \textbf{SIM} \(\uparrow\)} &
  { \textbf{NSS} \(\uparrow\)} \\ \hline
{ MDNSal \cite{reddy2020tidying}}  & { 0.217} & { 0.899} & { 0.797} & { 1.893} \\
{ SimpleNet \cite{reddy2020tidying}} &
  { 0.193} &
  { 0.907} &
  { 0.797} &
  { 1.926} \\
{ EML-NET \cite{jia2020eml}} & { \textbf{0.204}} & { 0.890} & { 0.785} & { 2.024} \\
{ \textbf{RecSal-NIS}}    & { 0.206} & { \textbf{0.907}} & { \textbf{0.803}} & { \textbf{2.027}} \\ \hline
\end{tabular}

}
\vspace{-1pt}
\caption{\footnotesize Comparison with prior art in SALICON validation dataset}
\label{SALICONval}
}
\hfill
\parbox{.49\linewidth}{\resizebox{0.49\columnwidth}{!}{
\centering
\begin{tabular}{l|llll}
\hline
{ } &
  \multicolumn{4}{c}{{ \textbf{MIT1003}}} \\
\multirow{-2}{*}{{ \textbf{Method}}} &
  \multicolumn{1}{c}{{ \textbf{KL} \(\downarrow\)}} &
  \multicolumn{1}{c}{{ \textbf{CC} \(\uparrow\)}} &
  \multicolumn{1}{c}{{ \textbf{SIM} \(\uparrow\)}} &
  \multicolumn{1}{c}{{ \textbf{NSS} \(\uparrow\)}} \\ \hline
{ DPNSal \cite{oyama2018influence}} &
  { \textbf{0.368}} &
  { 0.692} &
  { \textbf{0.813}} &
  { 2.678} \\
{ DeepFix \cite{kruthiventi2017deepfix}} &
  { -} &
  { 0.720} &
  { 0.540} &
  { 2.580} \\
{ SAM-VGG \cite{cornia2018predicting}} &
  { -} &
  { 0.757} &
  { -} &
  { 2.852} \\
{ SAM-ResNet \cite{cornia2018predicting}} &
  { -} &
  { 0.768} &
  { -} &
  { 2.893} \\
{ \textbf{RecSal-NIS}} &
  { 0.665} &
  { \textbf{0.784}} &
  { 0.583} &
  { \textbf{3.074}} \\ \hline
\end{tabular}}
\vspace{0.5pt}
\caption{\footnotesize Comparison with prior art in MIT1003 validation dataset}
\label{MITval}
}
\end{table}

%%%%%

\textbf{LSUN challenge 2017: \footnote{https://competitions.codalab.org/competitions/17136\#results}} We search the hyperparameter space again for improving on \(sAUC\) score since it is used for ranking in LSUN challenge. As it has been shown in \cite{tavakoli2017saliency} that \(NSS\) is very closely related to \(sAUC\), hence we choose to provide maximum weightage to \(L_{NSS}\) during training. The results of this training were submitted for the challenge where we secured the second position (Table \ref{LSUNtest}) which is commendable given the low parameter count of our model (Table \ref{parameter}). We consistently rank among top 5 in each metric with the exception of \(SIM\) where we stand at eighth position. Note that model trained with $\alpha, \beta, \gamma, \delta$ optimized for performing best on all the metrics (as in ablation study) when evaluated on the SALICON test set performs better for all other metrics but misses out on top 2 \(sAUC\) score.

\begin{table}[h]
\centering
\resizebox{0.9\columnwidth}{!}{
\begin{tabular}{l|ccccccc}
\hline
{ } &
  \multicolumn{7}{c}{{ \textbf{SALICON}}} \\
\multirow{-2}{*}{{ \textbf{Method}}} &
  { \textbf{sAUC} \(\uparrow\)} &
  { \textbf{IG} \(\uparrow\)} &
  { \textbf{NSS} \(\uparrow\)} &
  { \textbf{CC} \(\uparrow\)} &
  { \textbf{AUC} \(\uparrow\) } &
  { \textbf{SIM} \(\uparrow\)} &
  { \textbf{KL} \(\downarrow\)} \\ \hline
{ SimpleNet \cite{reddy2020tidying}} &
  { 0.743} &
  { 0.880} &
  { 1.960} &
  { 0.907} &
  { 0.869} &
  { 0.793} &
  { 0.201} \\
{ SAM-ResNet \cite{cornia2018predicting}} &
  { 0.741} &
  { 0.538} &
  { 1.990} &
  { 0.899} &
  { 0.865} &
  { 0.793} &
  { 0.610} \\
{ EML-NET \cite{jia2020eml}} &
  { 0.746} &
  { 0.736} &
  { 2.050} &
  { 0.886} &
  { 0.866} &
  { 0.780} &
  { 0.520} \\
{ MDNSal \cite{reddy2020tidying}} &
  { 0.736} &
  { 0.863} &
  { 1.935} &
  { 0.899} &
  { 0.865} &
  { 0.790} &
  { 0.221} \\
{ MD-SEM  \cite{fosco2019many}} &
  { 0.746} &
  { 0.660} &
  { 2.058} &
  { 0.868} &
  { 0.864} &
  { 0.774} &
  { 0.568} \\
{ \textbf{RecSal-NIS}} &
  { \textbf{0.747}} &
  { 0.854} &
  { 2.043} &
  { 0.900} &
  { 0.866} &
  { 0.789} &
  { 0.237} \\ \hline
\end{tabular}
}
\vspace{5pt}
\caption{\footnotesize Comparison with LSUN'17 leaderboard (ranking based on sAUC)
}
\label{LSUNtest}
\end{table}

We also attempted submitting our results for the MIT300 and CAT2000 saliency benchmark, but their servers are down. Hence we also validate our methods on the OSIE eye-tracking dataset and compared them with prior art in Table \ref{OSIEval}.

%%%%% do wala table
\begin{table}
\centering
% \hspace{0.5cm}
\parbox{.37\linewidth}{\resizebox{0.37\columnwidth}{!}{
\centering
\begin{tabular}{l|rc}
\hline
{ \textbf{Method}} & \multicolumn{1}{c}{{ \textbf{Parameters}}} & { \textbf{sAUC} \(\uparrow\)} \\ \hline
{ MD-SEM \cite{fosco2019many}}        & { 30.9 M}             & { 0.746} \\
{ SAM-ResNet \cite{cornia2018predicting}}    & { {\small$\sim$} 70 M}         & { 0.741} \\
{ EML-NET \cite{jia2020eml}}       & { {\small \textgreater} 100 M} & { 0.746} \\
{ SimpleNet \cite{reddy2020tidying}}     & { {\small \textgreater} 86 M}  & { 0.743} \\
{ \textbf{RecSal-NIS}} & { \textbf{15.56 M}}             & {\textbf{ 0.747}} \\ \hline
\end{tabular}
}
\vspace{-8pt}
\caption{\footnotesize Parameters vs sAUC comparison with prior art}
\label{parameter}
}
\hfill
\parbox{.61\linewidth}{\resizebox{0.61\columnwidth}{!}{
\centering
\begin{tabular}{l|cccc}
\hline
{ } &
  \multicolumn{4}{c}{{ \textbf{OSIE}}} \\
\multirow{-2}{*}{{ \textbf{Method}}} &
  { \textbf{KL} \(\downarrow\)} &
  { \textbf{CC} \(\uparrow\)} &
  { \textbf{SIM} \(\uparrow\)} &
  { \textbf{NSS} \(\uparrow\)} \\ \hline
 
{ SALICON (implemented by \cite{oyama2018influence})} &
  { 0.545} &
  { 0.605} &
  { 0.762} &
  { 2.762} \\
 
{ DenseSal \cite{oyama2018influence}} &
  { 0.443} &
  { 0.659} &
  { 0.822} &
  { 3.068} \\
{ DPNSal \cite{oyama2018influence}} &
  { 0.397} &
  { 0.686} &
  { \textbf{0.838}} &
  { 3.175} \\
{ \textbf{RecSal-NIS}} &
  { \textbf{0.326}} &
  { \textbf{0.864}} &
  { 0.720} &
  { \textbf{3.843}} \\ \hline
\end{tabular}
}\vspace{10pt}
\caption{\footnotesize Comparison with prior art on OSIE validation set}
\label{OSIEval}
}
\end{table}

%%%%%
% \vspace{-4pt}
\begin{figure*}
\includegraphics[width=\linewidth]{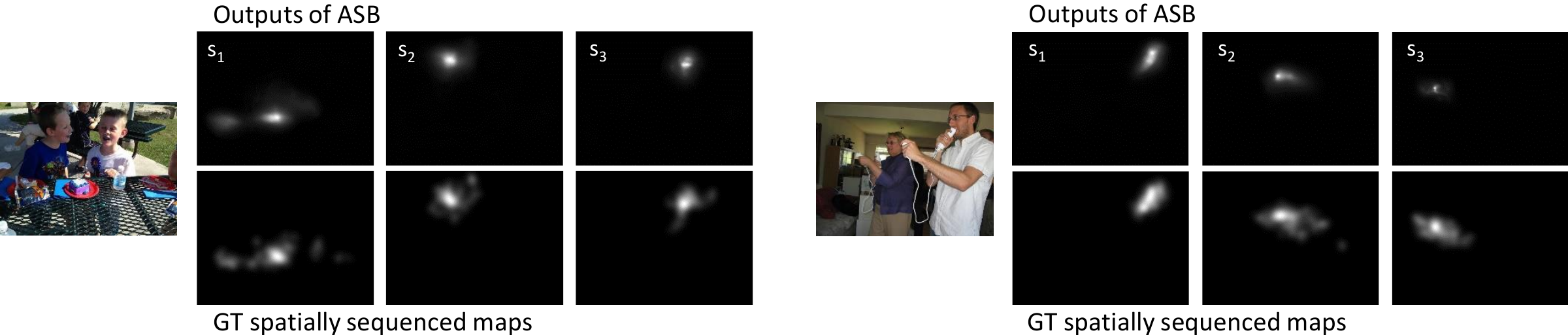}
\vspace{-10pt}
   \caption{\footnotesize Outputs produced by \(ASB\) of model supervised with non-incremental spatial data - RecSal-NIS. Images \(s_1, s_2\) and \(s_3\) are output maps of \(ASB\) after each iteration of \(RB\) in order of occurrence}
\label{fig:viz_spat}
\end{figure*}

% \begin{figure*}
% \includegraphics[width=\linewidth]{model_arch.pdf}
% \vspace{-5pt}
%   \caption{\footnotesize(a) Base encoder-multi decoder architecture, (b) Recursive Module for Incremental Data (\(RB\)), (c) Recursive Module for Non-Incremental Data built upon \(RB\) with the addition of \(HSAB\) module}
% \label{fig:arch}
% \end{figure*}

\section{Conclusion}
\label{sec:conclusion}

Experimental results demonstrate that applying recursive supervision using temporally and spatially sequenced data improves the performance over a given base model. We also find that the non-incrementally arranged spatial metadata method works better than all other variations. We believe it could be because separated spatial cues make it easier for the network to extract important features specific to those regions which contribute to the final saliency pattern. Our work suggests that improvement in performance does not necessarily require higher parameters, but rather an efficient usage of data.

% \newpage
\bibliography{egbib}
\end{document}